\pdfoutput=1

\documentclass[11pt]{article}

\usepackage[final]{acl}

\usepackage{times}
\usepackage{latexsym}

\usepackage[T1]{fontenc}

\usepackage[utf8]{inputenc}

\usepackage{microtype}

\usepackage{inconsolata}

\usepackage[inline]{enumitem}
\usepackage{hyperref}       
\usepackage{amsfonts}       
\usepackage{nicefrac}       
\usepackage{microtype}      
\usepackage{xcolor}         
\usepackage{booktabs}
\usepackage{amsmath}
\usepackage{multirow}
\usepackage{graphicx}
\usepackage{enumitem}
\usepackage{subcaption}
\usepackage{caption}
\usepackage{rotating}
\usepackage[normalem]{ulem}
\usepackage{amssymb}
\usepackage{colortbl}
\usepackage{linguex}

\usepackage{tikz,pgfplots,pgfplotstable}

\usepackage{siunitx} 
\usepackage{multirow, makecell}
\usepackage{pifont}
\usepackage{siunitx} 
\usepackage{multirow, makecell}
\usepackage{tabularx}
\usepackage[all]{nowidow}
\definecolor{darkgreen}{rgb}{0.0, 0.42, 0.24}
\definecolor{green}{RGB}{112, 173,71}
\definecolor{blue}{RGB}{68, 114,196}
\definecolor{orange}{RGB}{237, 125,49}
\definecolor{red}{RGB}{202, 54,49}
\definecolor{yellow}{RGB}{222,194, 142}

\newcommand{\bleurt}{\textsc{BLEURT}-20\xspace}
\newcommand{\bertscore}{\textsc{BertScore}\xspace}
\newcommand{\metricx}{\textsc{MetricX}\xspace}
\newcommand{\matese}{\textsc{MaTESe}\xspace}
\newcommand{\cometkiwi}{\textsc{CometKiwi}\xspace}
\newcommand{\nogroup}{\textsc{No-Grouping}\xspace}
\newcommand{\srcgroup}{\textsc{Group-by-Src}\xspace}

\definecolor{darkblue}{rgb}{0,0,.5}
\definecolor{darkgreen}{rgb}{0,.5,0}
\definecolor{lightgray}{rgb}{.8,.8,.8}
\definecolor{aliceblue}{rgb}{0.75, 0.75, 1.0}
\definecolor{darkseagreen}{rgb}{0.46, 0.74, 0.46}
\definecolor{alizarin}{rgb}{0.82, 0.1, 0.26}
\definecolor{airforceblue}{rgb}{0.36, 0.54, 0.66}
\definecolor{red_graph}{rgb}{0.98, 0.8, 0.8}
\definecolor{blue_graph}{rgb}{0.8, 0.98, 0.8}
\definecolor{red}{rgb}{0.8, 0.0, 0.0}

\newcommand{\cmss}[1]{{\fontfamily{cmss}\selectfont{#1}}}

\newcommand{\dataviz}[3]{\begin{tikzpicture}[scale=1.5]
\filldraw[draw=alizarin,fill=alizarin!30] (#2,0) rectangle (#1+#2+#3,0.15); 
\filldraw[draw=darkseagreen, fill=darkseagreen!30] (#1,0) rectangle (#1+#2,0.15);
\filldraw[draw=darkgreen, fill=darkgreen!30] (0,0) rectangle (#1,0.15);
\end{tikzpicture}}

\usepackage{graphicx}

\usepackage{hyperref}
\usepackage{comment}
\usepackage{graphicx}
\usepackage{multirow}

\usepackage{arydshln}
\usepackage{makecell}
\makeatletter
\def\adl@drawiv#1#2#3{%
        \hskip.5\tabcolsep
        \xleaders#3{#2.5\@tempdimb #1{1}#2.5\@tempdimb}%
                #2\z@ plus1fil minus1fil\relax
        \hskip.5\tabcolsep}
\newcommand{\cdashlinelr}[1]{%
  \noalign{\vskip 1.3pt
           \global\let\@dashdrawstore\adl@draw
           \global\let\adl@draw\adl@drawiv}
  \cdashline{#1}[.4pt/2pt]
  \noalign{\global\let\adl@draw\@dashdrawstore
           \vskip 1.3pt}}
\makeatother

\makeatletter
\newcommand{\printfnsymbol}[1]{%
  \textsuperscript{\@fnsymbol{#1}}%
}
\makeatother

\title{
Can Automatic Metrics Assess High-Quality Translations? }

\author{Sweta Agrawal$^{1}$\thanks{Equal contribution.}, António Farinhas$^{1,2*}$, \textbf{Ricardo Rei}$^{3}$, \textbf{André F.T. Martins}$^{1,2,3,4}$ \\
$^1$Instituto de Telecomunicações, $^2$Instituto Superior Técnico, Universidade de Lisboa\\
$^3$Unbabel, $^4$ELLIS Unit Lisbon\\
\small\texttt{swetaagrawal20@gmail.com, antonio.farinhas@tecnico.ulisboa.pt}}

\begin{document}
\maketitle
\begin{abstract}

Automatic metrics for evaluating translation quality are typically validated by measuring how well they correlate with human assessments. 
However, correlation methods tend to capture only the ability of metrics to differentiate between good and bad source-translation pairs, overlooking their reliability in distinguishing alternative translations for the same source. 
In this paper, we confirm that this is indeed the case by showing that current metrics are insensitive to nuanced differences in translation quality. This effect is most pronounced when the quality is high and the variance among alternatives is low. 
Given this finding, we shift towards detecting high-quality correct translations, an important problem in practical decision-making scenarios where a binary check of correctness is prioritized over a nuanced evaluation of quality. Using the MQM framework as the gold standard, we systematically stress-test the ability of current metrics to identify translations with no errors as marked by humans. Our findings reveal that current metrics often over or underestimate translation quality, indicating significant room for improvement in machine translation evaluation.

\end{abstract}

\section{Introduction}

The automatic evaluation of machine or human-generated translations has gained widespread attention over the past few years. 
Evaluation metrics act as proxies for translation quality in the absence of human judgments, offering immediate feedback.
They are widely used not only to provide quality indicators to users and translators \cite{informatics8030061, castilho2017acceptability, mehandru-etal-2023-physician}, but also to improve machine translation (MT) systems themselves \citep{he2024improving, xu2024a, fernandes-etal-2022-quality}.

\begin{table}[!t]
    \centering
    \scalebox{0.90}{
    \begin{tabular}{rlrcc}
     & \textsc{lp} & \multicolumn{1}{c}{\textsc{N}}  & \multicolumn{1}{c}{\textsc{\% zero-mqm}}  & \\
    \rowcolor{gray!10}
    \multicolumn{5}{c}{\textsc{WMT 2023 metrics dataset}} \\
    & \textsc{En-De (P)} & $5520$ & $25.4$\% &  \dataviz{0.254}{0.272}{0.474} \\
    & \textsc{He-En} & $9840$&  $50.8$\%  &  \dataviz{0.508}{0.193}{0.299}\\
    & \textsc{Zh-En} & $17655$& $19.1$\% & \dataviz{0.191}{0.456}{0.353} \\
    \addlinespace[0.5em]
    \rowcolor{gray!10}
    \multicolumn{5}{c}{\textsc{WMT 2022 metrics dataset}} \\
     & \textsc{En-De} &  $18410$  &  $51.5$\%& \dataviz{0.515}{0.356}{0.129} \\
     & \textsc{En-Ru}   & $19725$&  $42.7$\% & \dataviz{0.427}{0.312}{0.261} \\
     & \textsc{Zh-En}  & $26250$ &  $46.4$\% & \dataviz{0.464}{0.191}{0.345} \\
     \addlinespace[0.5em]
      \rowcolor{gray!10}
    \multicolumn{5}{c}{\textsc{WMT 2022 chat dataset}} \\
     & \textsc{XX-En} &  $4756$ & $63.2$\% &  \dataviz{0.632}{0.171}{0.197} \\
     & \textsc{En-XX}  & $5901$ & $60.2$\% &  \dataviz{0.602}{0.194}{0.205} \\
    \end{tabular}
    }
    \caption{Gold MQM scores distribution in recent WMT datasets. High-quality translations are represented in shades of green (darker for MQM $=0$ and lighter for MQM $\geq -5$); red represents translations with at least one major error (MQM $\leq-5$). P: paragraph-level.
    }
    \label{tab:general_stats}
\end{table}

Judging whether, and to what extent, these metrics concur with human evaluation is important to ensuring their effectiveness and applicability in diverse scenarios.
A recent human evaluation study by the Conference on Machine Translation (WMT)  revealed that translations produced by current MT systems often achieve very high-quality scores (ranging from $80$ to $90$) when judged by humans on a direct assessment (DA) scale of $0$ to $100$ \citep{kocmi-etal-2023-findings}.
Similarly, \citet{deutsch-etal-2023-ties} observe that these systems increasingly generate numerous ``perfect'' translations (translations with zero errors), especially for high-resource language pairs, as shown in Table~\ref{tab:general_stats}. 
As MT quality advances, evaluating whether evaluation metrics accurately reflect this progress is essential \cite{burchardt2016towards}. The absence of clear criteria for assessing these high-quality translations can introduce bias, leading to inconsistent assessments based on metric preferences rather than objective measures of accuracy.

Most evaluations of automatic metrics primarily assess their ability to distinguish between good and bad source-translation pairs \citep{freitag-etal-2023-results, freitag-etal-2022-results}, often overlooking their capacity to discern subtle differences in 
quality for a given source.
Furthermore, in many practical and high-risk applications (\textit{e.g.}, within the medical or legal domains), the main concern is not 
measuring the \textit{accuracy level} of a translation but determining whether \textit{the translation is accurate and fit for that specific use} 
\cite{nida1964toward, church1993good, bowker2019fit, vieira2021understanding, mehandru2023physician}.
While correlations provide valuable insights into the performance of automatic metrics, they do not offer a definitive measure of whether these metrics can reliably confirm translation accuracy. 

Hence, in this work, we systematically investigate how existing MT metrics assess high-quality (HQ) correct translations, defined as translations with zero or minor errors only. We find that automatic metrics struggle to distinguish between translations for a given source, especially when comparing HQ translations, with reference-free or quality estimation (QE) metrics achieving close correlation scores to reference-based ones. We further show that current metrics severely overestimate (for non-HQ translations) or underestimate (for HQ translations) translation quality. \cmss{GEMBA-MQM} \citep{kocmi-federmann-2023-gemba}, a GPT-based QE metric, achieves the highest F1 score in detecting the HQ translations with no errors (\textsc{HQ-Zero}). However, it also assigns high scores to erroneous GPT-4 translations, suggesting a preferential bias towards the LLM's own outputs. These findings highlight the necessity for more robust evaluation protocols to assess the quality of automatic metrics.


\section{How good are current MT systems?}

%
The most reliable way to assess translation quality has been through human evaluations, with several frameworks proposed over the years for this purpose.
While earlier works consider two dimensions---adequacy and fluency---with a 5-point Likert scale \cite{king1996evaluating}, subsequent work on direct assessments (DA) considers a single continuous scale of $0-100$ \citep{graham2017can}. However, several studies have questioned the credibility of DA-based evaluation \cite{toral-etal-2018-attaining, laubli2020set, fischer-laubli-2020-whats, mathur-etal-2020-results, freitag-etal-2021-experts}.

Unlike DAs, which assign a numeric score to a translation, the recent Multidimensional Quality Metrics \citep[MQM]{burchardt-2013-multidimensional} framework relies on explicit error judgments (error types and severities) marked within specific spans of the source-translation pair, providing a more accurate and fine-grained evaluation.
Translations receive a score of $0$ if they contain no errors, a penalty of $-1$ for minor errors, and $-5$ for major errors that impact the usage or understanding of the content.\footnote{Although the MQM framework includes critical errors---errors that could render a text unusable---they are not marked in many datasets due to their highly contextual interpretation.}

We present the distribution of gold MQM scores from the WMT23 Metrics task \cite{freitag-etal-2023-results}, WMT22 Metrics task \cite{freitag-etal-2022-results}, and WMT22 Chat Translation task \cite{farinha-etal-2022-findings} in Table~\ref{tab:general_stats}. Across settings and language pairs, the percentage of translations achieving a zero MQM score ranges from $19.1$\% to $63.2$\%. At least $52.6$\% of the translations across language pairs and settings have no major errors (MQM $>-5$).
Thus, a large percentage of the datasets include translations with no or only minor errors, emphasizing the importance of accurately identifying these high-quality translations in the evaluation process.

\section{How well do MT metrics assess HQ translations?}

We define HQ translations as those that achieve an MQM score $>-5$, \textit{i.e.}, translations without any major errors according to human evaluators. By definition, these translations 
do not contain errors that impede their comprehension or usability. We consider a subset of QE and reference-based automatic metrics evaluated by the shared tasks (see App.~\ref{sec:metrics} for more details).

\subsection{How do metrics rank HQ translations?}

We first investigate how automatic metrics rank HQ translations, 
which is particularly relevant today, as these metrics are often used to guide MT training or decoding processes. 
Recent work employs both reference-based and QE metrics to rerank multiple hypotheses generated by dedicated MT models or large language models (LLMs), aiming to improve translation quality \citep{fernandes-etal-2022-quality, freitag-etal-2022-high, farinhas-etal-2023-empirical, farinhas2024rerankinglawslanguagegeneration}. These metrics are also used to provide quality feedback signals during training, either explicitly in loss signals \cite{ramos2023aligning, yan-etal-2023-bleurt, he2024improving} or implicitly via the creation of preference datasets \cite{xu2024contrastive, yang2023direct}.

\begin{figure*}
    \centering
    \includegraphics[width=0.90\linewidth, trim={1cm 2.3cm 0.5cm 2cm},clip]{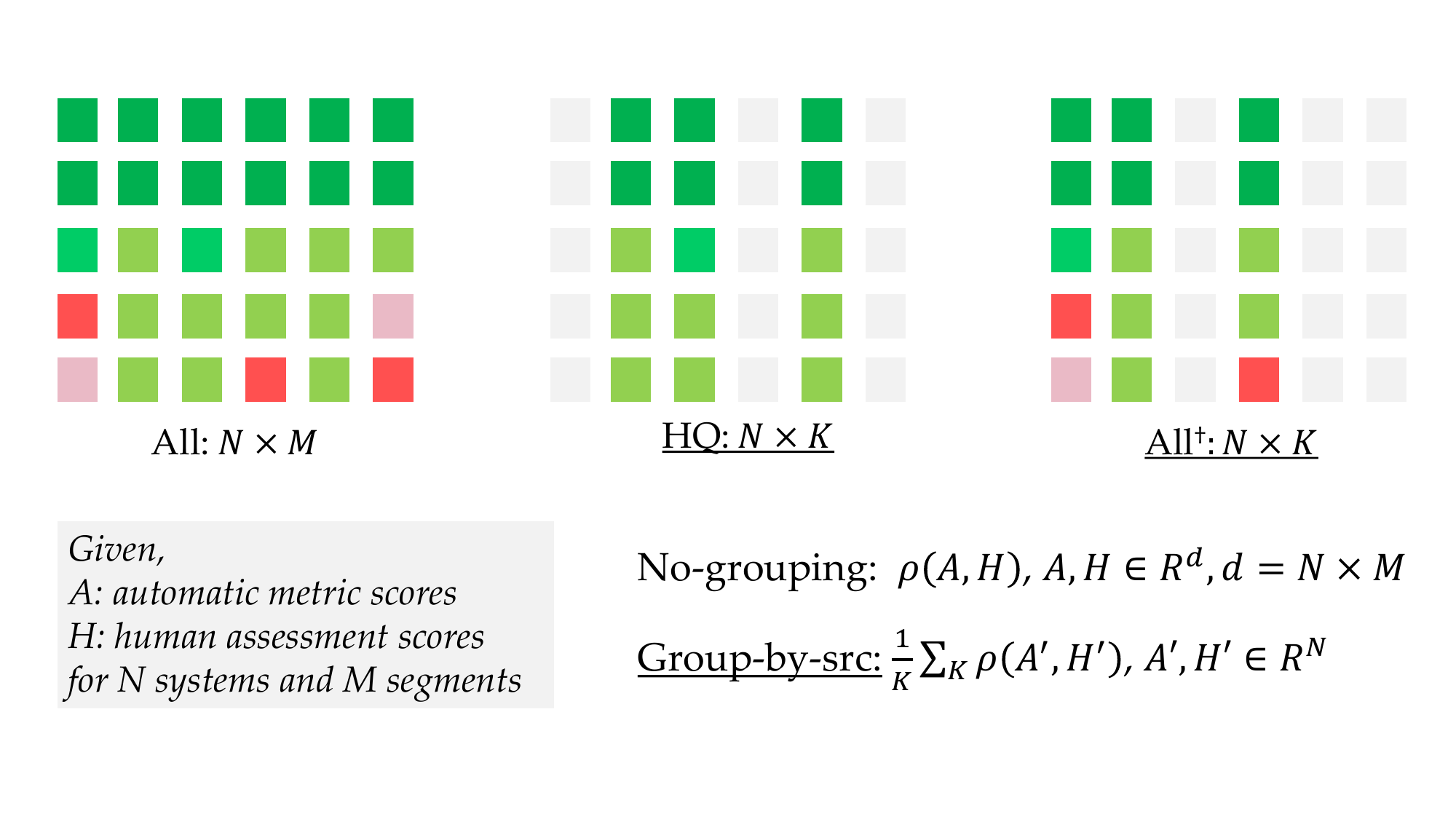}
    \caption{Ranking analysis configurations. $\rho$: Spearman correlation.}
    \label{fig:correlation}
\end{figure*}

Consider $N$ systems and $M$ source segments. Typically, segment-level correlations are computed between the $N\times M$ translations. 
However, this differs from the practical setting where metrics are used to rerank several translations for the same source. 
Therefore, we follow \citet{deutsch-etal-2023-ties} and compute the average correlation between the $N$ translation scores grouped by the source sentences. We refer to the former setting as \textbf{\nogroup} and the latter as \textbf{\srcgroup}.
We also study to what extent these metrics distinguish between HQ translations. 
As the number of segments with all HQ translations, $K$, is less than $M$, we report mean correlations on subsampled datasets (randomly sampled $10$ times) that match the sample size, $N \times K$, marked with the symbol $\dagger$ in Table~\ref{table:hq-np-wmt23-ende}. 
This is motivated by \citet{mathur-etal-2020-tangled}, who study how these metrics rank HQ \textbf{systems}, where a limited number of samples (typically 4 or 5) was shown to yield unreliable conclusions.
However, our focus is on \textbf{segment-level} evaluation, where the number of subsampled items is much larger. Figure~\ref{fig:correlation} summarizes all configurations and the corresponding correlation measures. 

Table~\ref{table:hq-np-wmt23-ende} presents Spearman correlation of automatic metrics with MQM scores for configurations described above on the WMT23 EN-DE dataset (see App.~\ref{sec:other_hq_np} for other datasets and correlation metrics).
We first note that the correlation observed on the entire (\nogroup ALL) and the subsampled datasets (\nogroup ALL$^\dagger$) is close, 
establishing that the observed differences cannot be merely attributed to changes in sample size.


\begin{table}[t]
\centering
\setlength\tabcolsep{3pt}
\scalebox{0.8}{
\begin{tabular}{llcc@{\hskip 0.2in}cc}
\toprule
& & \multicolumn{2}{c}{\nogroup}  & \multicolumn{2}{c}{\srcgroup}
\\
\cmidrule(lr){3-4} \cmidrule(lr){5-6} 
 & \multirow{-2}{*}{\rotatebox{0}{{\textsc{Metric}}}}& ALL & ALL$^\dagger$  & ALL$^\dagger$ & HQ  \\
\midrule
\multirow{9}{*}{\rotatebox[origin=c]{90}{\textsc{Ref-based}}} & \cmss{chrF}  &  $0.262$  &  $0.227$  & $0.267$  & $0.136$ \\
& \cmss{BLEU}  &  $0.193$ &  $0.190$   & $0.303$ & $0.146$  \\
& \cmss{BERTscore}  &  $0.355$  &  $0.367$  & $0.325$ & $0.134$ \\
& \cmss{COMET}  &  $0.578$  &  $0.584$  & $0.461$ & $0.202$  \\
& \cmss{BLEURT-20}  &  $0.618$ &  $0.603$ & $0.449$ & $0.220$ \\
& \cmss{XCOMET-XL}  &  $0.713$  &  $0.705$   & $0.461$  & $0.250$ \\
& \cmss{XCOMET-XXL}  &  $0.708$  &  $0.716$  & $0.481$  & $0.326$   \\
& \cmss{MetricX-23}  &  $0.682$  &  $0.680$ & $0.450$ & $0.301$   \\
& \cmss{MaTESe}  &  $0.591$ &  $0.593$  & $0.341$  & $0.254$  \\

\midrule
\multirow{5}{*}{\rotatebox[origin=c]{90}{\textsc{Ref-Free}}} & \cmss{GEMBA-MQM}  &  $0.614$  &  $0.621$ & $0.462$ & $0.368$  \\
& \cmss{CometKiwi}  &  $0.565$ &  $0.561$  & $0.411$ & $0.182$   \\
& \cmss{CometKiwi-XL}  &  $0.542$  &  $0.550$  & $0.427$  & $0.223$   \\
& \cmss{CometKiwi-XXL}  &  $0.525$ &  $0.504$  & $0.456$ & $0.327$  \\
& \cmss{MetricX-23-QE}  &  $0.683$ & $0.681$ & $0.470$ & $0.292$  \\
\bottomrule
\end{tabular}
}
\caption{Spearman correlation on WMT23 EN-DE. 
$\dagger$: Subsampled to match \srcgroup HQ's  size.
}\label{table:hq-np-wmt23-ende}
\end{table}

\begin{figure*}[h] 
\centering
 \begin{subfigure}{.32\textwidth}
  \includegraphics[width=0.98\linewidth ]{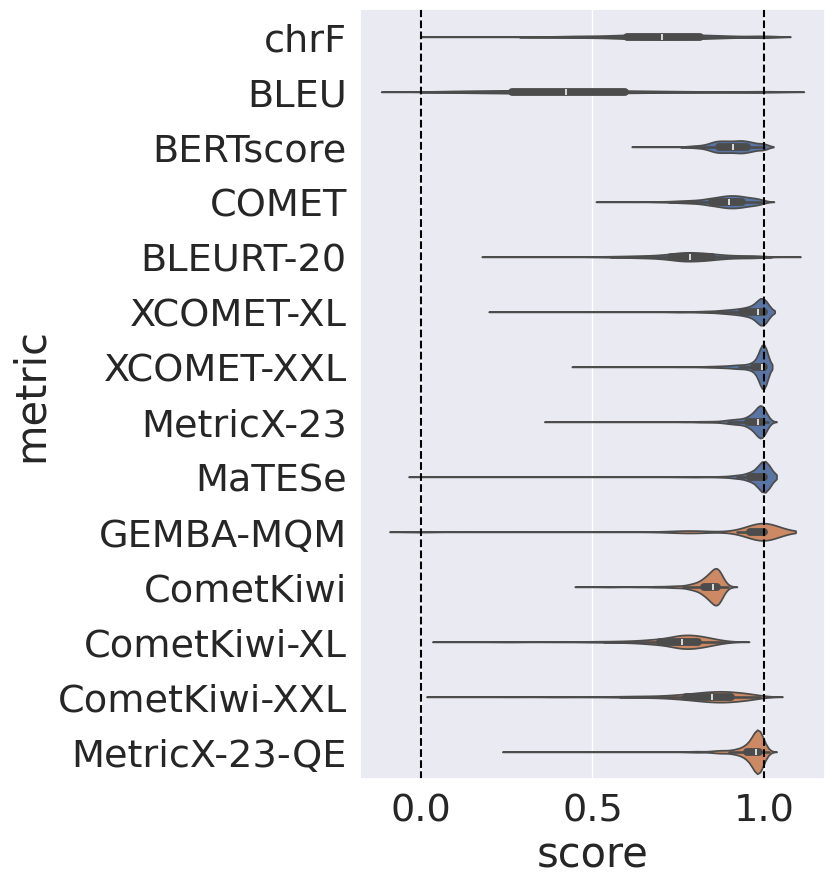}
     \end{subfigure}
\begin{subfigure}{.32\textwidth}
  \includegraphics[width=0.98\linewidth ]{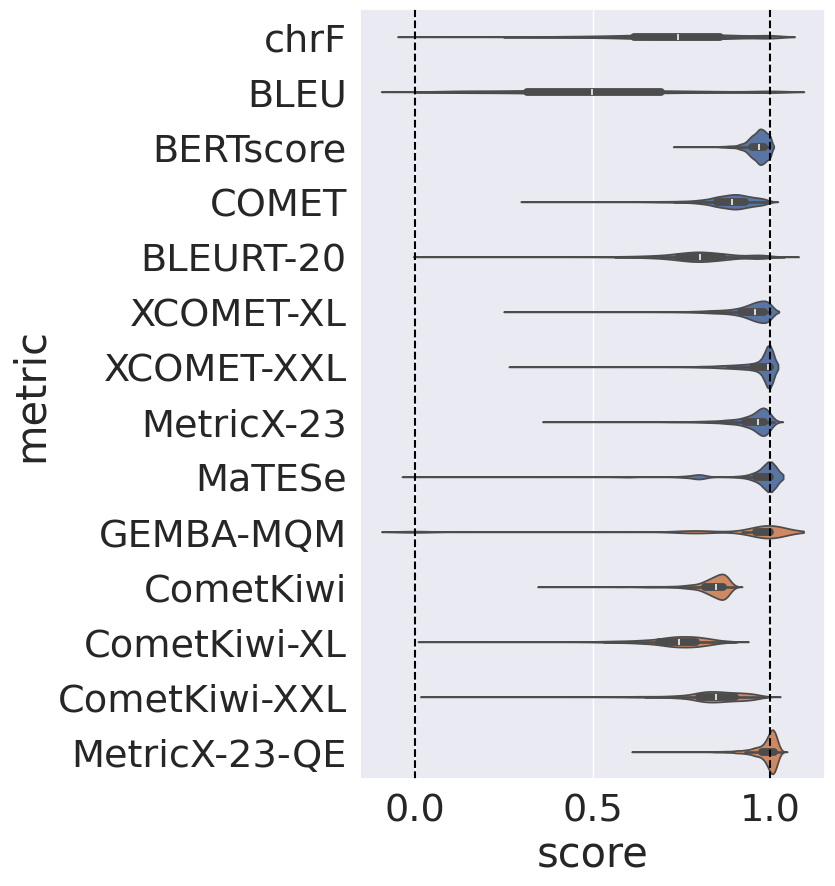}
  \end{subfigure}
     \begin{subfigure}{.32\textwidth}
  \includegraphics[width=0.98\linewidth ]{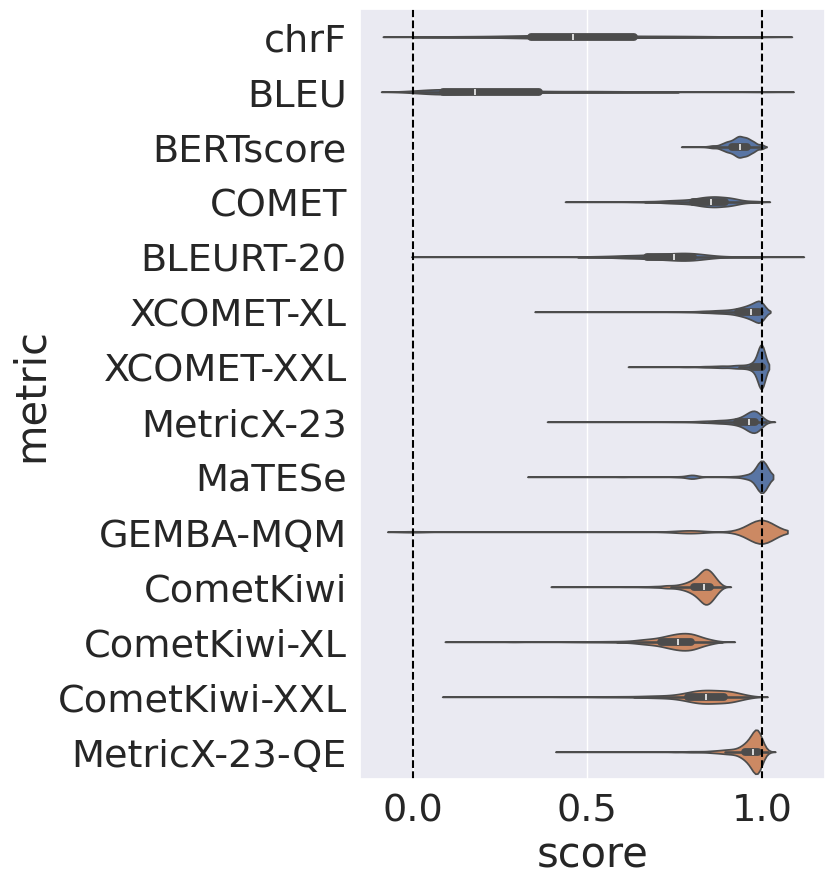}
     \end{subfigure}

     \bigskip
     
     \begin{subfigure}{0.9\linewidth }
        \centering
        \setlength\tabcolsep{8pt}
        
    \begin{tabular}{lccccccccccc}
        \rowcolor{gray!10}
        & \multicolumn{3}{c}{\textbf{\textsc{EN-DE ($1402$)}}} && \multicolumn{3}{c}{\textbf{\textsc{HE-EN ($5001$)}}}&& \multicolumn{3}{c}{\textbf{\textsc{ZH-EN ($11309$)}}} \\
        \rowcolor{gray!10}
        \multirow{-2}{*}{\textbf{\textsc{METRIC}}} & \textbf{\textsc{P}} & \textbf{\textsc{R}} &  \textbf{\textsc{F1}} && \textbf{\textsc{P}} & \textbf{\textsc{R}} &  \textbf{\textsc{F1}} && \textbf{\textsc{P}} & \textbf{\textsc{R}} &  \textbf{\textsc{F1}}   \\
        \cmss{xCOMET-XL} & $\textbf{72}$ & $40$ & $51$ & & $\textbf{78}$ & $17$ & $28$&& $47$ & $28$ & $35$\\   
        \cmss{xCOMET-XXL}   & $58$ & $59$ & $58$  & & $74$ & $54$ & $62$ && $36$ & $63$ & $46$ \\  
        \cmss{MaTESe}   &  $49$ & $69$ & $58$  & & $66$ & $\textbf{65}$ & $65$&&  $29$ & $\textbf{75}$ & $42$    \\
        \cmss{MetricX-23}	&  $70$ & $33$ & $45$ && $80$ & $16$ & $27$  &&  $52$ & $11$ & $19$  \\
        \cmss{GEMBA-MQM}   &$52$ & $\textbf{70}$ & $\textbf{60}$  & & $71$ & $\textbf{65}$ & $\textbf{68}$&&  $37$ & $77$ & $\textbf{50}$\\ 
        \cmss{MetricX-23-QE}	&  $66$ & $14$ & $23$ &&  $70$ & $64$ & $67$ && $\textbf{55}$ & $20$ & $29$  \\
    \end{tabular} 
    \end{subfigure}
        \caption{\textbf{Top:} Metric Scores distribution for HQ-\textsc{Zero} translations on WMT23. \textbf{Bottom:} Precision, recall, and F1. }  \label{table:hq_p_wmt23} 
\end{figure*}

\noindent
\paragraph{Metrics exhibit only a low-to-fair correlation with human judgments when evaluating translations for the same source. }
Automatic metrics 
are less effective in differentiating between good and bad translations for the same source, as evidenced by 
the drop in correlation from the \nogroup \textsc{ALL}$^\dagger$ to the \srcgroup \textsc{ALL}$^\dagger$ setting. 
A possible reason for this disparity lies in how these metrics are typically trained---most metrics are trained to predict translation quality for a given instance (\textit{e.g.}, source-reference-hypothesis trio in \cmss{Comet} or \cmss{xCOMET}). 
While useful for ranking two \textit{systems} based on averaged scores across texts, they may provide limited information for gauging translation quality for different translations of the same source.\footnote{Using contrastive objectives or exposing the metric to multiple translations could potentially help mitigate this issue \cite{briakou-carpuat-2020-detecting}.} Interestingly, \cmss{BLEU}'s correlation is higher in the \srcgroup setting than in \nogroup, likely due to its original use for comparing multiple translations of the same source.
This underscores the limitations of using automatic metrics as the sole measure of quality beyond their intended use cases, particularly in scenarios where fine-grained distinctions between translations of the same source 
are required.

\noindent
\paragraph{QE metrics are on par with reference-based ones for differentiating translations. } 
QE metrics show promising results in differentiating translations 
for the same source, often achieving comparable or better correlation than reference-based metrics. 
For EN-DE, the QE metrics
\cmss{\textsf{MetricX-23-QE}} and \cmss{\textsf{GEMBA-MQM}} rank second and third, respectively
in the \textsc{ALL} setting, following \cmss{xCOMET-XXL}. When contrasting HQ translations, \cmss{\textsf{GEMBA-MQM}} outperforms all other metrics. The relatively strong performance of QE metrics, particularly in this setting, highlights their potential as valuable tools for translation generation and ranking tasks.

\noindent
\paragraph{Metrics fail to distinguish HQ translations.}

There is a consistent drop in correlation scores across all metrics in the HQ relative to the ALL setting, possibly because most translations in the HQ setting receive scores in the narrow range of $(-5, 0]$ and are often tied in quality.
\citet{deutsch-etal-2023-ties} show that most metrics struggle to predict translation ties accurately, \textit{i.e.}, to give the same score to two translations with similar quality, except for error-predicting metrics like \cmss{GEMBA-MQM} or \cmss{MaTESe}. 
This decreased correlation from the HQ to the ALL setting has significant implications, especially when they are used to rerank translations produced by strong MT systems. It may result in an artificial boost or bias towards specific systems or outputs, inadvertently prioritizing translations that align well with metric biases but deviate from true quality improvements, as discussed in \S\ref{sec:bias}.

\subsection{How well do metrics detect HQ translations with no errors?}

Ranking translations of similar quality is a difficult task, so we also evaluate how automatic metrics score HQ translations with zero MQM scores. 
(HQ-\textsc{Zero}). We consider normalized scores $\ge 0.99$ as \textit{valid} scores as $1.0$ is the highest score a metric should assign to HQ-\textsc{Zero} translations.  Fig.~\ref{table:hq_p_wmt23} shows the results on WMT23 dataset.
See App.~\ref{sec:other_hq_p} for results in other datasets.

\paragraph{Metric scores have high variance for HQ translations.} $9$ out of $15$ metrics do not assign valid scores to HQ-\textsc{Zero} translations. 
Lexical metrics (\cmss{chrF} and \cmss{BLEU}) produce the lowest absolute values, possibly due to over-reliance on a reference translation. Neural metrics trained to regress on DA scores (\cmss{BLEURT}, \cmss{COMET}, and variants) also do not assign valid scores for these translations, likely due to low agreement between DA and MQM scores, as discussed by \citet{freitag-etal-2021-experts}. 

\paragraph{Metrics over or underestimate translation quality.} 
Metrics that do score these translations within the valid range (\cmss{xCOMET}, \cmss{MaTESe}, \cmss{MetricX}, and \cmss{GEMBA-MQM}), exhibit different tradeoffs between precision (P) and recall (R). For example, while \cmss{xCOMET-XL} and \cmss{MetricX} prioritize precision, \cmss{MaTESe} and \cmss{GEMBA-MQM} excel at recognizing many HQ-\textsc{Zero} translations, leading to increased recall.  
This difference might stem from the specific task each metric is optimized for: while the first two predict sentence-level quality, the last two are optimized to predict word-level error spans. 
As expected, \cmss{xCOMET-XXL} significantly outperforms \cmss{xCOMET-XL} across all language pairs.
Finally, the QE metric, \cmss{GEMBA-MQM}, based on GPT-4, achieves the highest F1 score across all language pairs, demonstrating the capabilities of LLM-based evaluation in more nuanced MT evaluation.

\subsection{Which HQ translations are detected?} \label{sec:bias}

\begin{figure}[t]
    \centering
    \includegraphics[width=0.99\linewidth]{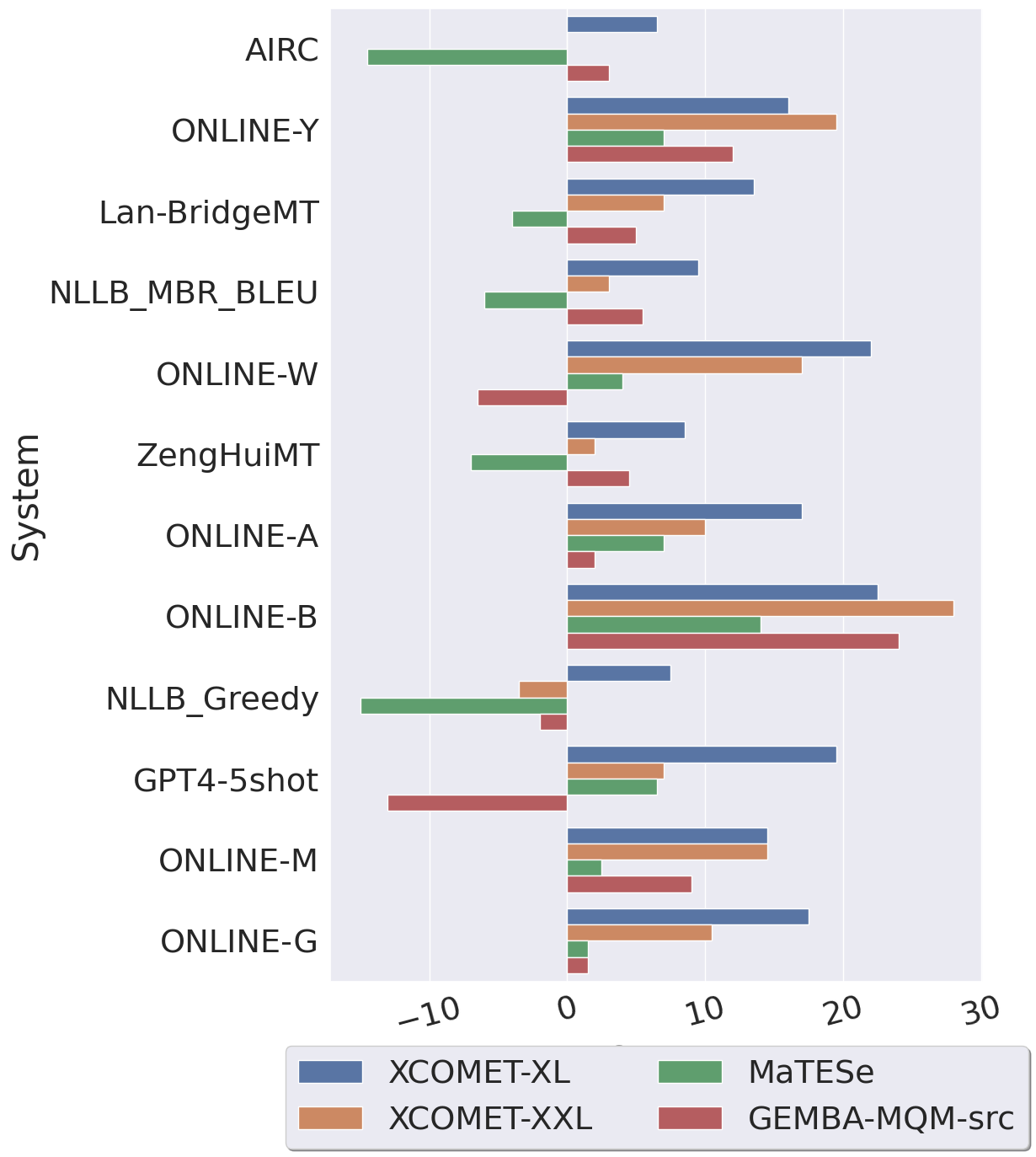}
    \caption{Absolute difference of the number of times a metric assigns a valid score to HQ-\textsc{Zero} and non HQ-\textsc{Zero} translations.}
    \label{fig:system_bias}
\end{figure}

To study preference bias from metrics towards specific systems, we compute the absolute difference in the number of times a metric assigns a valid score to HQ-\textsc{Zero} and non-HQ-\textsc{Zero} translations. Fig.~\ref{fig:system_bias} shows that \textsf{MaTESe} equally overestimates translation quality for many systems, as suggested by its high 
R and low 
P scores.
\cmss{GEMBA-MQM} frequently assigns zero MQM 
scores to GPT-4 
translations, even when humans identify errors in them.
This aligns with concurrent works showing a preference bias of LLMs towards their outputs \cite{panickssery2024llm, xu2024perils}, underscoring the need for a more detailed evaluation to better understand the outputs these metrics prefer and whether they align with human preferences.

\section{Conclusions and Future Work}

This work systematically investigates how automatic metrics assess HQ translations. We find that current metrics correlate poorly with human judgments when contrasting translations for a given source, with the correlation being even lower for HQ translations.
We then study whether metrics can detect HQ translations that attain zero MQM scores (HQ-\textsc{Zero}) and find that many metrics fail to assign them valid scores.
While the GPT-4-based \cmss{\textsf{GEMBA-MQM}} attains the highest F1 for detecting HQ-\textsc{Zero}, it shows some preference for GPT-4 outputs. 
Therefore, despite its promise, it is essential to complement \cmss{\textsf{GEMBA-MQM}} with other metrics to ensure robust evaluation.

\section*{Acknowledgments}

We thank Ben Peters, Gonçalo Faria, and Eleftheria Briakou for their constructive feedback on the paper.
This work was supported by EU's Horizon Europe Research and Innovation Actions (UTTER, contract 101070631), by the project DECOLLAGE (ERC-2022-CoG 101088763), by the Portuguese Recovery and Resilience Plan through project C645008882-00000055 (Center for Responsible AI), and by Fundação para a Ciência e Tecnologia through contract UIDB/50008/2020.

\section*{Limitations}
We highlight the main limitations of our work.
First, we rely on human MQM annotations as the gold standard for identifying high-quality translations, despite their potential subjectivity and occasional inaccuracy. These annotations are collected for individual translations, and the ratings might vary if annotators were asked to evaluate and compare multiple translations simultaneously. Furthermore, MQM annotations used in our analysis are very expensive to obtain as they require trained linguists to perform the assessments, which limits the analysis to publicly available datasets.

Second, although our analysis spans multiple datasets across six language pairs (\textsc{en-de}, \textsc{zh-en}, \textsc{he-en}, \textsc{en-ru}, \textsc{en-fr}, and \textsc{en-pt-br}) and multiple domains, we do not necessarily account for the distribution of high-quality translations across different domains within a dataset. As shown by \citet{zouhar2024fine}, learned metrics can be sensitive to the domain of evaluation.

Lastly, our analysis in \S\ref{sec:bias} identifies one potential bias, but it remains unclear whether automatic metrics have preferential biases towards other output properties such as length, stylistic choices, etc.



\bibliography{custom, anthology}

\appendix
\newpage
\onecolumn
\section{Automatic Metrics} \label{sec:metrics}

We present details about all automatic metrics used across different datasets in Table~\ref{tab:automatic}. We refer the reader to the relevant papers \cite{freitag-etal-2022-results, freitag-etal-2023-results, agrawal2024context} for more details.

We used the datasets and scores from the WMT 2022 and WMT 2023 Metrics Shared Task campaign, which are available at \url{https://github.com/google-research/mt-metrics-eval} under the Apache License Version 2.0. For WMT 2022 Chat Shared task human assessments, we used human assessments from \url{https://github.com/WMT-Chat-task/data-and-baselines/tree/main/data/mqm-annotations}, released under a CC-BY-NC license.  In our work, we ensured that our usage was consistent with their intended purposes as specified by the licenses.

\begin{table}[h]
\centering
\rotatebox[origin=c]{-90}{
\renewcommand\arraystretch{1.2}
\setlength\tabcolsep{4pt}
\scalebox{0.6}{
\begin{tabular}{rlllllp{5cm}p{5cm}}
\toprule
 \multicolumn{1}{c}{\textsc{\textbf{Metric}}} & \multicolumn{1}{c}{\textsc{\textbf{Paper}}}  & \multicolumn{1}{c}{\textsc{\textbf{Input}}}& \multicolumn{1}{c}{\textsc{\textbf{Output}}} & \multicolumn{1}{c}{\textsc{\textbf{Type}}} & \multicolumn{1}{c}{\textsc{\textbf{Evaluation}}} & \multicolumn{1}{c}{\textsc{\textbf{Dataset}}} & \multicolumn{1}{c}{\textsc{\textbf{Base Model}}} \\
\midrule
 \cmss{\textbf{chrF}} & \citet{popovic-2015-chrf} & \{\textsc{ref}, \textsc{mt}\} & [0-100] $\in \mathbb{R}$ & \textsc{lexical} & WMT22, WMT23 & - & - \\
\cmss{\textbf{BLEU}} & \citet{papineni-etal-2002-bleu} & \{\textsc{ref}, \textsc{mt}\} & [0-100] $\in \mathbb{R}$ & \textsc{lexical} & WMT22, WMT23 & - & - \\

\addlinespace[0.2cm]
\cmss{\textbf{\bertscore}} & \citet{Zhang2020BERTScore}& \{\textsc{ref}, \textsc{mt}\}  & {[0-1]} $\in \mathbb{R}$ & \textsc{Embedding} & WMT22, WMT23 & - & \texttt{bert-base-multilingual-cased} \\
\addlinespace[0.2cm]

\cmss{\textbf{COMET}} & \citet{rei-etal-2022-comet} &\{\textsc{src}, \textsc{ref}, \textsc{mt}\} &  [0-1] $\in \mathbb{R}$ 
 & \textsc{Learned} & WMT23 & DA (WMT 2017-2020) + MLQE-PE & \texttt{xlm-roberta-large} \\

\cmss{\textbf{\bleurt-20}} & \citet{sellam-etal-2020-bleurt}  &\{\textsc{ref}, \textsc{mt}\}  &  [0-1] $\in \mathbb{R}$
 & \textsc{Learned} & WMT22, WMT23 & DA (WMT 2015-2020) + Synthetic  & \texttt{rembert} \\

\cmss{\textbf{COMET-22}}$*$ & \citet{rei-etal-2022-comet} &\{\textsc{src}, \textsc{ref}, \textsc{mt}\}  &  [0-1] $\in \mathbb{R}$ & \textsc{Learned} & WMT22 & DA (WMT 2017-2020) + MLQE-PE  + MQM  & \texttt{xlm-roberta-large, infoxlm-large}  \\

\cmss{\textbf{\metricx-22}} & - &\{\textsc{ref}, \textsc{mt}\}  &  [-25,0], $\mathbb{R}$  & \textsc{Learned} & WMT22 & -  & \texttt{30B mT5}\\
\cmss{\textbf{\metricx-23}} & \citet{juraska-etal-2023-metricx} &\{\textsc{ref}, \textsc{mt}\}  &  [-25,0] $\in \mathbb{R}$  & \textsc{Learned} & WMT23 & DA (WMT 2015-2020) +  MQM (WMT 2020-2021) +  Synthetic &  \texttt{mT5-XXL} \\

\cmss{\textbf{xCOMET}}$*$ & \citet{guerreiro2023xcomet} &\{\textsc{src}, \textsc{ref}, \textsc{mt}\} & [0-1] $\in \mathbb{R}$ & \textsc{Learned} & WMT23 & DA (WMT 2017-2020) + MLQE-PE  + MQM  (WMT 2020-2021; IndicMT, DEMETR) + Synthetic  & \texttt{XLM-RoBERTa-XL, XLM-RoBERTa-XXL}  \\

\cmss{\textbf{\matese}} & \citet{perrella-etal-2022-matese}  & \{\textsc{ref}, \textsc{mt}\} & [-25,0]$\in \mathbb{Z}$ & textsc{Learned} & WMT22 & MQM (WMT 2020-2021) & \texttt{XLM-RoBERTa, BART}  \\

\cmss{\textbf{\matese}} & - & \{\textsc{ref}, \textsc{mt}\} & [-25,0]$\in \mathbb{Z}$ & textsc{Learned} & WMT23 & MQM (WMT 2020-2022) & \texttt{DeBERTa, InfoXLM}  \\

\addlinespace[0.2cm]
\cmss{\textbf{\textsc{GEMBA-MQM}}} & \citet{kocmi-federmann-2023-gemba} &\{\textsc{src}, \textsc{mt}\} & [-25,0] $\in \mathbb{Z}$ & LLM-based & WMT23 & - & \texttt{GPT4} \\

\cmss{\textbf{\cometkiwi-22}}$*$ & \citet{rei-etal-2022-cometkiwi} &\{\textsc{src}, \textsc{mt}\} &  [0-1] $\in \mathbb{R}$ & \textsc{Learned} & WMT22 & DA (WMT 2017-2020) + MLQE-PE  + MQM $^\dagger$  & \texttt{rembert, infoxlm-large}  \\
 
\cmss{\textbf{\cometkiwi-23}}$*$ & \citet{rei2023scaling} &\{\textsc{src}, \textsc{mt}\}   &  [0-1] $\in \mathbb{R}$ 
 & \textsc{Learned} & WMT23 & DA (WMT 2017-2020) + MLQE-PE  + MQM $^\dagger$  & \texttt{rembert, infoxlm-large}  \\
 
\cmss{\textbf{\metricx-23-QE}} &  \citet{juraska-etal-2023-metricx} &\{\textsc{src}, \textsc{mt}\}  &  [-25,0] $\in \mathbb{R}$  & \textsc{Learned} & WMT23 & DA (WMT 2015-2020) +  MQM (WMT 2020-2021) +  Synthetic &  \texttt{mT5-XXL} \\

\cmss{\textbf{\matese-QE}} & \citet{perrella-etal-2022-matese} &\{\textsc{src}, \textsc{mt}\} & [-25,0]$\in \mathbb{Z}$ & textsc{Learned} & WMT22 & MQM (WMT 2020-2021) & \texttt{XLM-RoBERTa, BART}  \\

\cmss{\textbf{xCOMET-QE}}$*$ & \citet{guerreiro2023xcomet}  &\{\textsc{src}, \textsc{mt}\} &  [0-1] $\in \mathbb{R}$ & \textsc{Learned} & WMT23 & DA (WMT 2017-2020) + MLQE-PE  + MQM  (WMT 2020-2021; IndicMT, DEMETR) + Synthetic  & \texttt{XLM-RoBERTa-XL, XLM-RoBERTa-XXL}  \\
\bottomrule
\end{tabular}
}
}
\caption{Details about the automatic metrics considered in our paper. $*$: submission is an ensemble; 
$^\dagger$: \{\textsc{src}, \textsc{ref}\} pairs are also added to the training data.\label{tab:automatic}}
\end{table}

\newpage
\section{Ranking results} \label{sec:other_hq_np}

Tables~\ref{table:app-hq-np-wmt23-ende} and \ref{table:hq-np-wmt23-ende-pear} report the Spearman and Pearson correlation results for WMT23 EN-DE, respectively. Tables~\ref{table:hq-np-wmt23-app} and~\ref{table:hq-np-wmt22-app} show the Spearman Correlation for the WMT22 and WMT23 datasets, respectively. We do not perform this analysis on chat data because the number of systems is $\le$ 5.


\begin{table*}[ht!]
\centering
\setlength\tabcolsep{3pt}
\scalebox{0.8}{
\begin{tabular}{lccc@{\hskip 0.2in}ccc@{\hskip 0.2in}ccc}
\toprule
& \multicolumn{2}{c}{\nogroup} && \multicolumn{2}{c}{\nogroup$\dagger$} && \multicolumn{2}{c}{\srcgroup} & \\
\cmidrule(lr){2-3} \cmidrule(lr){5-6} \cmidrule(lr){8-9}
\multirow{-2}{*}{\rotatebox{0}{{\textsc{Metric}}}}& ALL &HQ& $\Delta$ & ALL & HQ  & $\Delta$  & ALL$^\dagger$ & HQ  & $\Delta$ \\
\midrule
\cmss{chrF}  &  $0.262$ & $0.137$ & $-0.124$  &  $0.227$ \textcolor{violet}{\scriptsize$\pm 0.030$} &  $0.132$ \textcolor{violet}{\scriptsize$\pm 0.022$} & $-0.094$  & $0.267$ \textcolor{violet}{\scriptsize$\pm 0.050$} & $0.136$ & $-0.131$  \\
\cmss{BLEU}  &  $0.193$ & $0.094$ & $-0.099$  &  $0.190$ \textcolor{violet}{\scriptsize$\pm 0.032$} &  $0.087$ \textcolor{violet}{\scriptsize$\pm 0.022$} & $-0.103$  & $0.303$ \textcolor{violet}{\scriptsize$\pm 0.056$} & $0.146$ & $-0.156$  \\
\cmss{BERTscore}  &  $0.355$ & $0.190$ & $-0.165$  &  $0.367$ \textcolor{violet}{\scriptsize$\pm 0.039$} &  $0.183$ \textcolor{violet}{\scriptsize$\pm 0.032$} & $-0.184$  & $0.325$ \textcolor{violet}{\scriptsize$\pm 0.035$} & $0.134$ & $-0.191$  \\
\cmss{COMET}  &  $0.578$ & $0.385$ & $-0.194$  &  $0.584$ \textcolor{violet}{\scriptsize$\pm 0.024$} &  $0.390$ \textcolor{violet}{\scriptsize$\pm 0.031$} & $-0.194$  & $0.461$ \textcolor{violet}{\scriptsize$\pm 0.041$} & $0.202$ & $-0.259$  \\
\cmss{BLEURT-20}  &  $0.618$ & $0.357$ & $-0.262$  &  $0.603$ \textcolor{violet}{\scriptsize$\pm 0.020$} &  $0.357$ \textcolor{violet}{\scriptsize$\pm 0.033$} & $-0.246$  & $0.449$ \textcolor{violet}{\scriptsize$\pm 0.043$} & $0.220$ & $-0.229$  \\
\cmss{XCOMET-XL}  &  $0.713$ & $0.454$ & $-0.259$  &  $0.705$ \textcolor{violet}{\scriptsize$\pm 0.020$} &  $0.449$ \textcolor{violet}{\scriptsize$\pm 0.018$} & $-0.256$  & $0.461$ \textcolor{violet}{\scriptsize$\pm 0.030$} & $0.250$ & $-0.211$  \\
\cmss{XCOMET-XXL}  &  $0.708$ & $0.399$ & $-0.309$  &  $0.716$ \textcolor{violet}{\scriptsize$\pm 0.020$} &  $0.382$ \textcolor{violet}{\scriptsize$\pm 0.032$} & $-0.335$  & $0.481$ \textcolor{violet}{\scriptsize$\pm 0.041$} & $0.326$ & $-0.155$  \\
\cmss{MetricX-23}  &  $0.682$ & $0.433$ & $-0.249$  &  $0.680$ \textcolor{violet}{\scriptsize$\pm 0.018$} &  $0.446$ \textcolor{violet}{\scriptsize$\pm 0.027$} & $-0.233$  & $0.450$ \textcolor{violet}{\scriptsize$\pm 0.043$} & $0.301$ & $-0.149$  \\
\cmss{MaTESe}  &  $0.591$ & $0.353$ & $-0.238$  &  $0.593$ \textcolor{violet}{\scriptsize$\pm 0.028$} &  $0.370$ \textcolor{violet}{\scriptsize$\pm 0.044$} & $-0.223$  & $0.341$ \textcolor{violet}{\scriptsize$\pm 0.042$} & $0.254$ & $-0.087$  \\

\midrule
\multicolumn{10}{c}{\textit{quality estimation}}\\\cdashlinelr{1-10}
\addlinespace[0.2cm]
\cmss{GEMBA-MQM}  &  $0.614$ & $0.345$ & $-0.269$  &  $0.621$ \textcolor{violet}{\scriptsize$\pm 0.027$} &  $0.358$ \textcolor{violet}{\scriptsize$\pm 0.028$} & $-0.263$  & $0.462$ \textcolor{violet}{\scriptsize$\pm 0.044$} & $0.368$ & $-0.094$  \\
\cmss{CometKiwi}  &  $0.565$ & $0.286$ & $-0.279$  &  $0.561$ \textcolor{violet}{\scriptsize$\pm 0.019$} &  $0.268$ \textcolor{violet}{\scriptsize$\pm 0.021$} & $-0.293$  & $0.411$ \textcolor{violet}{\scriptsize$\pm 0.044$} & $0.182$ & $-0.229$  \\
\cmss{CometKiwi-XL}  &  $0.542$ & $0.240$ & $-0.302$  &  $0.550$ \textcolor{violet}{\scriptsize$\pm 0.023$} &  $0.254$ \textcolor{violet}{\scriptsize$\pm 0.032$} & $-0.296$  & $0.427$ \textcolor{violet}{\scriptsize$\pm 0.029$} & $0.223$ & $-0.204$  \\
\cmss{CometKiwi-XXL}  &  $0.525$ & $0.236$ & $-0.289$  &  $0.504$ \textcolor{violet}{\scriptsize$\pm 0.031$} &  $0.244$ \textcolor{violet}{\scriptsize$\pm 0.032$} & $-0.260$  & $0.456$ \textcolor{violet}{\scriptsize$\pm 0.029$} & $0.327$ & $-0.129$  \\
\cmss{MetricX-23-QE}  &  $0.683$ & $0.425$ & $-0.258$  &  $0.681$ \textcolor{violet}{\scriptsize$\pm 0.012$} &  $0.439$ \textcolor{violet}{\scriptsize$\pm 0.027$} & $-0.242$  & $0.470$ \textcolor{violet}{\scriptsize$\pm 0.028$} & $0.292$ & $-0.177$  \\
\bottomrule
\end{tabular}
}
\caption{Spearman correlation on WMT23 EN-DE. 
$\dagger$: Subsampled to match \srcgroup HQ's sample size.
}\label{table:app-hq-np-wmt23-ende}
\end{table*}

\begin{table*}[ht!]
\centering
\setlength\tabcolsep{3pt}
\scalebox{0.8}{
\begin{tabular}{lccc@{\hskip 0.2in}ccc@{\hskip 0.2in}ccc}
\toprule
& \multicolumn{2}{c}{\nogroup} && \multicolumn{2}{c}{\nogroup$\dagger$} && \multicolumn{2}{c}{\srcgroup} & \\
\cmidrule(lr){2-3} \cmidrule(lr){5-6} \cmidrule(lr){8-9}
\multirow{-2}{*}{\rotatebox{0}{{\textsc{Metric}}}}& ALL &HQ& $\Delta$ & ALL & HQ  & $\Delta$  & ALL$^\dagger$ & HQ  & $\Delta$ \\
\midrule
\cmss{chrF}  &  $0.232$ & $0.112$ & $-0.120$  &  $0.244$ \textcolor{violet}{\scriptsize$\pm 0.028$} &  $0.121$ \textcolor{violet}{\scriptsize$\pm 0.028$} & $-0.123$  & $0.322$ \textcolor{violet}{\scriptsize$\pm 0.041$} & $0.124$ & $-0.198$  \\
\cmss{BLEU}  &  $0.192$ & $0.086$ & $-0.106$  &  $0.210$ \textcolor{violet}{\scriptsize$\pm 0.029$} &  $0.079$ \textcolor{violet}{\scriptsize$\pm 0.025$} & $-0.131$  & $0.297$ \textcolor{violet}{\scriptsize$\pm 0.049$} & $0.148$ & $-0.149$  \\
\cmss{BERTscore}  &  $0.325$ & $0.150$ & $-0.175$  &  $0.331$ \textcolor{violet}{\scriptsize$\pm 0.038$} &  $0.148$ \textcolor{violet}{\scriptsize$\pm 0.031$} & $-0.182$  & $0.363$ \textcolor{violet}{\scriptsize$\pm 0.043$} & $0.150$ & $-0.213$  \\
\cmss{COMET}  &  $0.432$ & $0.337$ & $-0.095$  &  $0.421$ \textcolor{violet}{\scriptsize$\pm 0.037$} &  $0.367$ \textcolor{violet}{\scriptsize$\pm 0.031$} & $-0.055$  & $0.513$ \textcolor{violet}{\scriptsize$\pm 0.044$} & $0.266$ & $-0.246$  \\
\cmss{BLEURT-20}  &  $0.484$ & $0.324$ & $-0.160$  &  $0.488$ \textcolor{violet}{\scriptsize$\pm 0.021$} &  $0.308$ \textcolor{violet}{\scriptsize$\pm 0.024$} & $-0.180$  & $0.469$ \textcolor{violet}{\scriptsize$\pm 0.047$} & $0.245$ & $-0.223$  \\
\cmss{XCOMET-XL}  &  $0.680$ & $0.414$ & $-0.266$  &  $0.680$ \textcolor{violet}{\scriptsize$\pm 0.028$} &  $0.409$ \textcolor{violet}{\scriptsize$\pm 0.040$} & $-0.272$  & $0.510$ \textcolor{violet}{\scriptsize$\pm 0.054$} & $0.359$ & $-0.150$  \\
\cmss{XCOMET-XXL}  &  $0.695$ & $0.362$ & $-0.333$  &  $0.688$ \textcolor{violet}{\scriptsize$\pm 0.019$} &  $0.355$ \textcolor{violet}{\scriptsize$\pm 0.038$} & $-0.333$  & $0.484$ \textcolor{violet}{\scriptsize$\pm 0.068$} & $0.385$ & $-0.098$  \\
\cmss{MetricX-23}  &  $0.585$ & $0.406$ & $-0.179$  &  $0.576$ \textcolor{violet}{\scriptsize$\pm 0.023$} &  $0.406$ \textcolor{violet}{\scriptsize$\pm 0.025$} & $-0.169$  & $0.512$ \textcolor{violet}{\scriptsize$\pm 0.024$} & $0.371$ & $-0.141$  \\
\cmss{MaTESe}  &  $0.554$ & $0.238$ & $-0.316$  &  $0.547$ \textcolor{violet}{\scriptsize$\pm 0.035$} &  $0.221$ \textcolor{violet}{\scriptsize$\pm 0.032$} & $-0.325$  & $0.345$ \textcolor{violet}{\scriptsize$\pm 0.045$} & $0.253$ & $-0.092$  \\

\midrule
\multicolumn{10}{c}{\textit{quality estimation}}\\\cdashlinelr{1-10}
\addlinespace[0.2cm]
\cmss{GEMBA-MQM}  &  $0.502$ & $0.223$ & $-0.279$  &  $0.497$ \textcolor{violet}{\scriptsize$\pm 0.027$} &  $0.238$ \textcolor{violet}{\scriptsize$\pm 0.021$} & $-0.260$  & $0.485$ \textcolor{violet}{\scriptsize$\pm 0.055$} & $0.386$ & $-0.099$  \\
\cmss{CometKiwi}  &  $0.475$ & $0.210$ & $-0.265$  &  $0.476$ \textcolor{violet}{\scriptsize$\pm 0.037$} &  $0.198$ \textcolor{violet}{\scriptsize$\pm 0.049$} & $-0.277$  & $0.458$ \textcolor{violet}{\scriptsize$\pm 0.057$} & $0.226$ & $-0.232$  \\
\cmss{CometKiwi-XL}  &  $0.446$ & $0.185$ & $-0.262$  &  $0.445$ \textcolor{violet}{\scriptsize$\pm 0.033$} &  $0.198$ \textcolor{violet}{\scriptsize$\pm 0.032$} & $-0.247$  & $0.499$ \textcolor{violet}{\scriptsize$\pm 0.041$} & $0.328$ & $-0.171$  \\
\cmss{CometKiwi-XXL}  &  $0.417$ & $0.171$ & $-0.245$  &  $0.411$ \textcolor{violet}{\scriptsize$\pm 0.024$} &  $0.167$ \textcolor{violet}{\scriptsize$\pm 0.040$} & $-0.244$  & $0.531$ \textcolor{violet}{\scriptsize$\pm 0.040$} & $0.378$ & $-0.152$  \\
\cmss{MetricX-23-QE}  &  $0.626$ & $0.371$ & $-0.255$  &  $0.640$ \textcolor{violet}{\scriptsize$\pm 0.036$} &  $0.372$ \textcolor{violet}{\scriptsize$\pm 0.029$} & $-0.268$  & $0.536$ \textcolor{violet}{\scriptsize$\pm 0.048$} & $0.407$ & $-0.129$  \\

\bottomrule
\end{tabular}
}
\caption{Pearson correlation on WMT23 EN-DE. $\dagger$: Subsampled to match \srcgroup HQ's sample size.
}\label{table:hq-np-wmt23-ende-pear}
\end{table*}


\begin{table*}[ht!]

\small
\centering
\resizebox{1.0\linewidth}{!}{
\begin{tabular}{lrrrrrrrr}
\toprule
 & \multicolumn{4}{c}{WMT23 HE-EN} & \multicolumn{4}{c}{WMT23 ZH-EN} \\
\cmidrule(lr){2-5} \cmidrule(lr){6-9}
& \multicolumn{2}{c}{\nogroup$^\dagger$}& \multicolumn{2}{c}{\srcgroup} & \multicolumn{2}{c}{\nogroup$^\dagger$}& \multicolumn{2}{c}{\srcgroup}  \\
\cmidrule(lr){2-3} \cmidrule(lr){4-5} \cmidrule(lr){6-7} \cmidrule(lr){8-9}
\textbf{\textsc{Metric}}& \multicolumn{1}{c}{All} & \multicolumn{1}{c}{HQ}& \multicolumn{1}{c}{All$^\dagger$} & \multicolumn{1}{c}{HQ}& \multicolumn{1}{c}{All} & \multicolumn{1}{c}{HQ}& \multicolumn{1}{c}{All$^\dagger$} & \multicolumn{1}{c}{HQ}  \\
\midrule

\cmss{chrF}  &  $0.299$  & $0.140$  & $0.298$  & $0.144$  &  $0.067$  & $0.012$  & $0.220$  & $0.162$  \\
\cmss{BLEU}  &  $0.248$  & $0.145$  & $0.270$  & $0.161$  &  $0.129$  & $0.065$  & $0.190$  & $0.139$  \\
\cmss{BERTscore}  &  $0.391$  & $0.210$  & $0.368$  & $0.191$ &  $0.269$  & $0.129$  & $0.273$  & $0.154$  \\
\cmss{COMET}  &  $0.485$  & $0.226$  & $0.383$  & $0.167$  &  $0.457$  & $0.268$  & $0.315$  & $0.183$  \\
\cmss{BLEURT-20}  &  $0.459$  & $0.216$  & $0.379$  & $0.173$ &  $0.434$  & $0.241$  & $0.332$  & $0.189$  \\
\cmss{XCOMET-XL}  &  $0.511$  & $0.255$  & $0.362$  & $0.147$  &  $0.608$  & $0.405$  & $0.334$  & $0.185$  \\
\cmss{XCOMET-XXL}  &  $0.528$  & $0.260$  & $0.381$  & $0.140$ &  $0.607$  & $0.364$  & $0.373$  & $0.219$  \\
\cmss{MetricX-23}  &  $0.549$  & $0.258$  & $0.357$  & $0.171$ &  $0.603$  & $0.408$  & $0.339$  & $0.202$  \\
\cmss{MaTESe}  &  $0.415$  & $0.207$  & $0.353$  & $0.266$ &  $0.467$  & $0.277$  & $0.322$  & $0.216$  \\

\midrule
\multicolumn{9}{c}{\textit{quality estimation}}\\\cdashlinelr{1-9}
\addlinespace[0.2cm]
\cmss{GEMBA-MQM}  &  $0.493$  & $0.245$  & $0.420$  & $0.227$ &  $0.580$  & $0.358$  & $0.423$  & $0.264$  \\
\cmss{CometKiwi}  &  $0.459$  & $0.225$  & $0.309$  & $0.106$ &  $0.533$  & $0.328$  & $0.333$  & $0.160$  \\
\cmss{CometKiwi-XL}  &  $0.434$  & $0.184$  & $0.348$  & $0.181$  &  $0.532$  & $0.302$  & $0.334$  & $0.170$  \\
\cmss{CometKiwi-XXL}  &  $0.468$  & $0.213$  & $0.389$  & $0.202$  &  $0.504$  & $0.288$  & $0.352$  & $0.161$  \\
\cmss{MetricX-23-QE}  &  $0.495$  & $0.235$  & $0.307$  & $0.126$  &  $0.621$  & $0.411$  & $0.322$  & $0.159$  \\
\cmss{XCOMET-QE-Ensemble}  &  $0.504$  & $0.233$  & $0.345$  & $0.160$ &  $0.631$  & $0.377$  & $0.347$  & $0.177$  \\

\bottomrule
\end{tabular}
}
\caption{Spearman correlation on WMT23 (HE-EN and ZH-EN). $\dagger$: Subsampled to match \srcgroup HQ's sample size.}\label{table:hq-np-wmt23-app} 
\end{table*}

\begin{table*}[ht!]

\small
\centering
\resizebox{1.0\linewidth}{!}{
\setlength\tabcolsep{4pt}
\begin{tabular}{lrrrrrrrrrrrr}
\toprule
 & \multicolumn{4}{c}{WMT22 EN-DE} & \multicolumn{4}{c}{WMT22 EN-RU}  & \multicolumn{4}{c}{WMT22 ZH-EN}\\
\cmidrule(lr){2-5} \cmidrule(lr){6-9}  \cmidrule(lr){10-13}
& \multicolumn{2}{c}{\nogroup$^\dagger$}& \multicolumn{2}{c}{\srcgroup} & \multicolumn{2}{c}{\nogroup$^\dagger$}& \multicolumn{2}{c}{\srcgroup}& \multicolumn{2}{c}{\nogroup$^\dagger$}& \multicolumn{2}{c}{\srcgroup}  \\
\cmidrule(lr){2-3} \cmidrule(lr){4-5} \cmidrule(lr){6-7} \cmidrule(lr){8-9} \cmidrule(lr){10-11} \cmidrule(lr){12-13}
\textbf{\textsc{Metric}}& 
\multicolumn{1}{c}{All} & \multicolumn{1}{c}{HQ}&
\multicolumn{1}{c}{All$^\dagger$} & \multicolumn{1}{c}{HQ}& 
\multicolumn{1}{c}{All} & \multicolumn{1}{c}{HQ}& 
\multicolumn{1}{c}{All$^\dagger$} & \multicolumn{1}{c}{HQ} & 
\multicolumn{1}{c}{All} & \multicolumn{1}{c}{HQ} & 
\multicolumn{1}{c}{All$^\dagger$} & \multicolumn{1}{c}{HQ} \\
\midrule
\cmss{chrF}  &  $0.296$  & $0.214$  & $0.242$  & $0.206$&  $0.235$  & $0.161$  & $0.237$  & $0.161$   &  $0.199$  & $0.069$  & $0.189$  & $0.096$  \\
\cmss{BLEU}  &  $0.233$  & $0.176$  & $0.221$  & $0.210$  &  $0.194$  & $0.161$  & $0.198$  & $0.127$&  $0.200$  & $0.086$  & $0.146$  & $0.089$  \\
\cmss{BERTScore}  &  $0.318$  & $0.244$  & $0.239$  & $0.207$  &  $0.265$  & $0.210$  & $0.240$  & $0.158$ &  $0.428$  & $0.189$  & $0.265$  & $0.155$  \\
\cmss{COMET-22}  &  $0.497$  & $0.392$  & $0.358$  & $0.314$  &  $0.534$  & $0.387$  & $0.394$  & $0.282$ &  $0.428$  & $0.189$  & $0.265$  & $0.155$  \\
\cmss{BLEURT-20}  &  $0.467$  & $0.346$  & $0.352$  & $0.283$ &  $0.483$  & $0.342$  & $0.354$  & $0.257$ &  $0.488$  & $0.194$  & $0.305$  & $0.170$  \\
\cmss{MetricX-XL}  &  $0.499$  & $0.379$  & $0.395$  & $0.349$ &  $0.511$  & $0.392$  & $0.379$  & $0.290$ &  $0.550$  & $0.253$  & $0.314$  & $0.210$  \\
\cmss{MetricX-XXL}  &  $0.490$  & $0.377$  & $0.370$  & $0.304$ &  $0.561$  & $0.430$  & $0.402$  & $0.338$  &  $0.554$  & $0.260$  & $0.303$  & $0.204$ \\
\cmss{MaTESe}  &  $0.387$  & $0.296$  & $0.356$  & $0.349$ &  $0.315$  & $0.236$  & $0.321$  & $0.281$ &  $0.477$  & $0.243$  & $0.251$  & $0.222$  \\
\midrule
\multicolumn{13}{c}{\textit{quality estimation}}\\\cdashlinelr{1-13}
\addlinespace[0.2cm]
\cmss{CometKiwi}  &  $0.404$  & $0.300$  & $0.273$  & $0.223$&  $0.482$  & $0.341$  & $0.306$  & $0.228$ &  $0.488$  & $0.223$  & $0.263$  & $0.205$   \\
\cmss{MaTESe-QE}  &  $0.294$  & $0.236$  & $0.314$  & $0.316$ &  $0.258$  & $0.184$  & $0.268$  & $0.256$  &  $0.412$  & $0.214$  & $0.212$  & $0.208$  \\
\bottomrule
\end{tabular}
}
\caption{Spearman correlation on WMT22 (EN-DE, EN-RU, annd ZH-EN). $^\dagger$: Subsampled to match \srcgroup HQ's sample size.} \label{table:hq-np-wmt22-app}
\end{table*}

\newpage

\section{HQ-\textsc{Zero} Detection Results} \label{sec:other_hq_p}

\begin{figure}[h] 
\centering
 \begin{subfigure}{.32\columnwidth}
  \includegraphics[width=0.98\linewidth ]{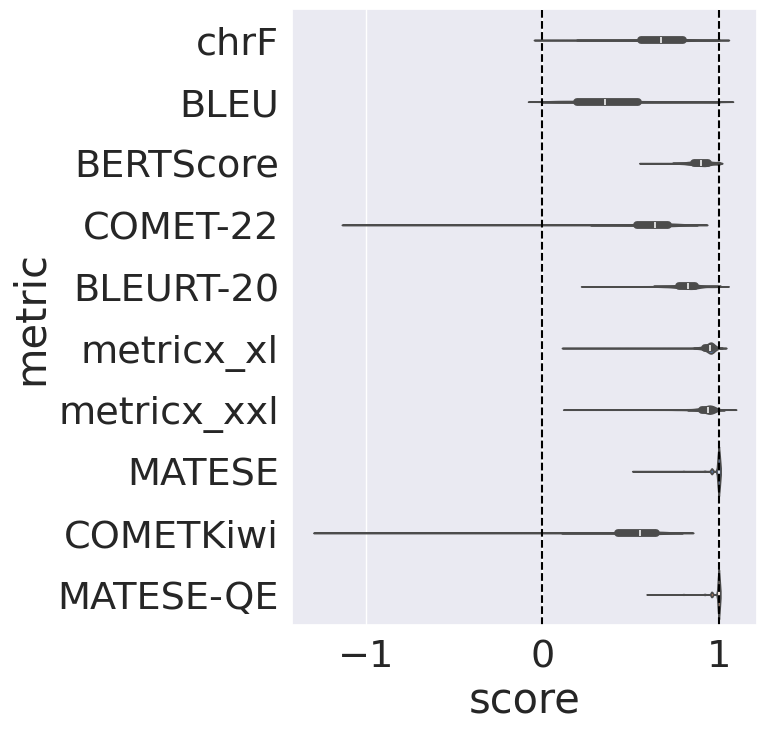}
     \end{subfigure}
\begin{subfigure}{.32\columnwidth}
  \includegraphics[width=0.98\linewidth ]{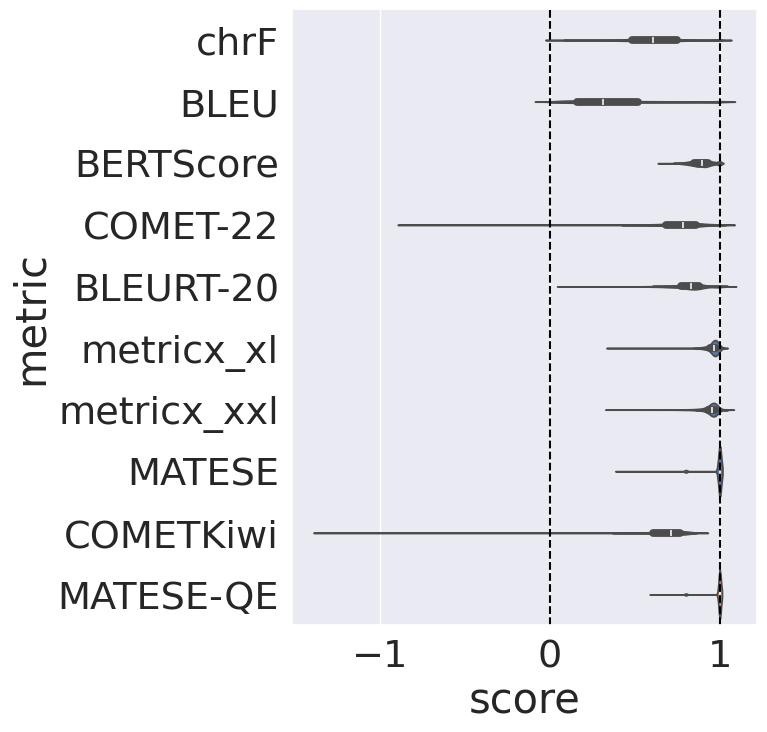}
  \end{subfigure}
     \begin{subfigure}{.32\columnwidth}
  \includegraphics[width=0.98\linewidth ]{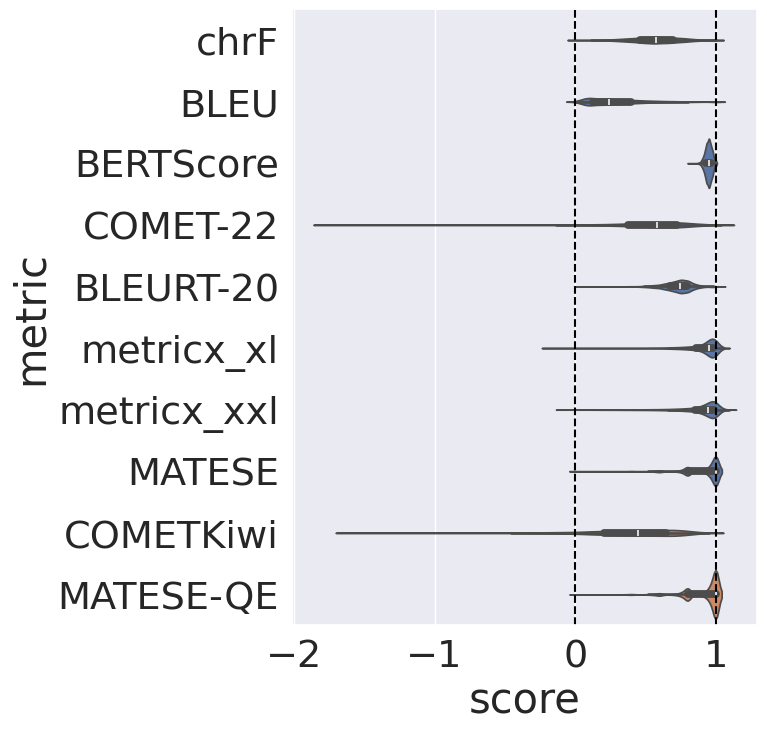}
     \end{subfigure}
     
     \bigskip
     
     \begin{subfigure}{0.9\linewidth }
        \centering
        \setlength\tabcolsep{8pt}
    \begin{tabular}{lccccccccccc}
    \rowcolor{gray!10}
 & \multicolumn{3}{c}{\textbf{\textsc{EN-DE}}} && \multicolumn{3}{c}{\textbf{\textsc{EN-RU}}}&& \multicolumn{3}{c}{\textbf{\textsc{ZH-EN}}} \\
\rowcolor{gray!10}
\multirow{-2}{*}{\textbf{\textsc{METRIC}}} & \textbf{\textsc{P}} & \textbf{\textsc{R}} &  \textbf{\textsc{F1}} && \textbf{\textsc{P}} & \textbf{\textsc{R}} &  \textbf{\textsc{F1}} && \textbf{\textsc{P}} & \textbf{\textsc{R}} &  \textbf{\textsc{F1}}   \\
\cmss{MaTESe}&$61$ & $86$ & $71$ && 	$48$ & $94$ & $63$ && $68$ & $53$ & $60$ 	\\
\cmss{MaTESe-QE} &  $58$ & $87$ & $70$ && $46$ & $95$ & $62$ &&  $64$ & $55$ & $59$\\
    \end{tabular} 
    \end{subfigure}
        \caption{\textbf{Top:} Scores distribution for HQ-\textsc{Zero} translations on WMT22. \textbf{Bottom:} Precision, recall, and F1.}  \label{table:hq_p_wmt22} 
\end{figure}

\begin{figure}[h] 
\centering
 \begin{subfigure}{.45\columnwidth}
  \includegraphics[width=0.98\linewidth ]{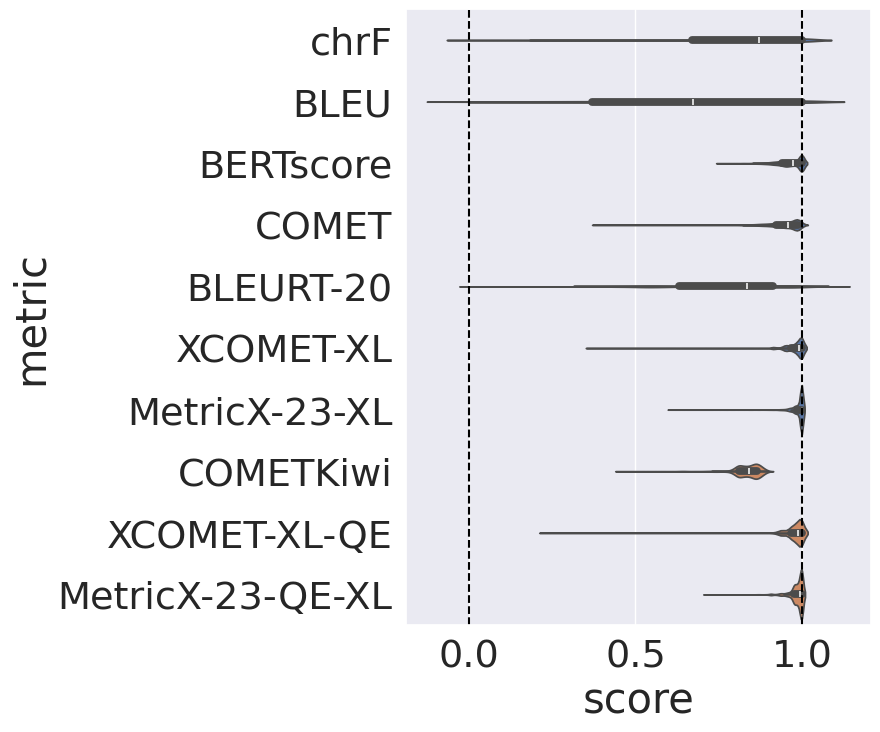}
     \end{subfigure}
\begin{subfigure}{.45\columnwidth}
  \includegraphics[width=0.98\linewidth ]{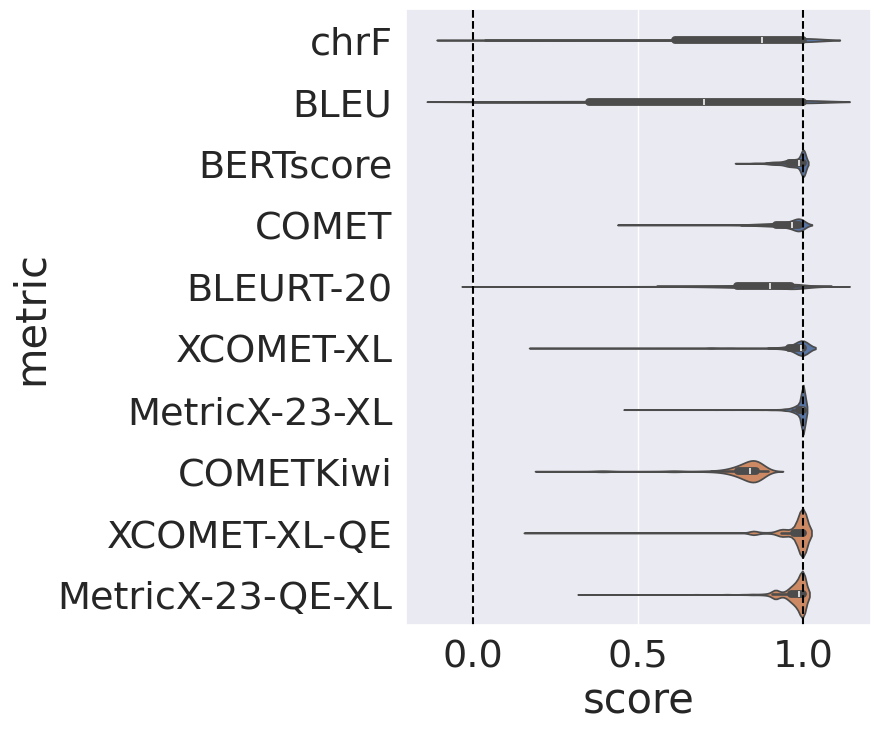}
  \end{subfigure}
     \bigskip
     \begin{subfigure}{0.9\linewidth }
        \centering
        \setlength\tabcolsep{8pt}
    \begin{tabular}{lccccccc}
    \rowcolor{gray!10}
 & \multicolumn{3}{c}{\textbf{\textsc{EN-XX}}} && \multicolumn{3}{c}{\textbf{\textsc{XX-EN}}}\\
\rowcolor{gray!10}
\multirow{-2}{*}{\textbf{\textsc{METRIC}}} & \textbf{\textsc{P}} & \textbf{\textsc{R}} &  \textbf{\textsc{F1}} && \textbf{\textsc{P}} & \textbf{\textsc{R}} &  \textbf{\textsc{F1}}    \\
\cmss{chrF}& $88$ & $38$ & $53$&& $92$ & $42$ & $58$  \\
\cmss{BLEU}& $88$ & $38$ & $53$&& $93$ & $42$ & $58$  \\
\cmss{BERTScore} & $93$ & $23$ & $37$&& $94$ & $27$ & $42$  \\
\cmss{XCOMET-XL}& $75$ & $33$ & $46$&& $87$ & $38$ & $53$  \\
\cmss{MetricX-23-XL} & $76$ & $64$ & $69$&& $87$ & $62$ & $72$  \\
\addlinespace[0.1cm]
\cmss{XCOMET-XL-QE}  & $66$ & $29$ & $40$&& $84$ & $49$ & $62$  \\
\cmss{MetricX-23-QE-XL}   & $76$ & $45$ & $56$&& $80$ & $35$ & $49$  \\
    \end{tabular} 
    \end{subfigure}
        \caption{\textbf{Top:} Scores distribution for HQ-\textsc{Zero} translations on WMT22 Chat. \textbf{Bottom:} Precision, recall, and F1.}  \label{table:hq_p_wmt22_chat} 
\end{figure}

We present the results for the detection task on the WMT22 Metrics and Chat datasets in Figures~\ref{table:hq_p_wmt22} and~\ref{table:hq_p_wmt22_chat}, respectively.

\end{document}